\documentclass{INGUADY}

\usepackage{makeidx}  % allows for indexgeneration
\usepackage[utf8]{inputenc}
\usepackage[T1]{fontenc}
\usepackage{cite}
\usepackage{amsfonts}
\usepackage{amsmath}
\usepackage{siunitx}
\usepackage{booktabs}

\begin{document}

% The title of your article
\title{Un modelo variacional nuevo para la clasificación binaria en el contexto de aprendizaje automático supervisado}

% The authors of your paper
\author {
	\correspauthor{C. D. Brito Pacheco} \orcidID{0000-0002-5913-2847},
	C. F. Brito Loeza
}

% The running head of your article
\runhead{C.D. Brito Pacheco \textit{et al.} / Ingeniería 22-1 (2018) 9-18}

% The front matter section of your article
\begin{frontmatter}

% These are the review dates that will be displayed
\reviewdates{30 de noviembre de 2017}{8 de febrero de 2018}

% The spanish abstract!
\begin{abstract}[Resumen]
Se examina el problema de aprendizaje supervisado en su planteación continua. Posteriormente se da una condición de optimalidad general a través de técnicas del análisis funcional y el cálculo de variaciones. Esto nos permite resolver la condición de optimalidad para la función deseada $u$ numéricamente y hacer varias comparaciones con otros modelos de aprendizaje supervisado ampliamente utilizados. Se emplea la precisión porcentual y el área bajo la curva característica operativa del receptor (AUC por sus siglas en inglés) como métricas del rendimiento. Finalmente, se realizan 3 análisis basados en estas dos métricas mencionadas donde comparamos los modelos y hacemos conclusiones para determinar si nuestro método es competitivo o no.

% Spanish keywords
\keywords[Palabras Clave]{aprendizaje variacional, aprendizaje automático, inteligencia artificial}
\end{abstract}

% English title of your publication
\englishtitle{A New Variational Model for Binary Classification in the Supervised Learning Context}

% English abstract
\begin{abstract}[Abstract]
We examine the supervised learning problem in its continuous setting and give a general optimality condition through techniques of functional analysis and the calculus of variations. This enables us to solve the optimality condition for the desired function $u$ numerically and make several comparisons with other widely utilized supervised learning models. We employ the accuracy and area under the receiver operating characteristic curve as metrics of the performance. Finally 3 analyses are conducted based on these two mentioned metrics where we compare the models and make conclusions to determine whether  or not our method is competitive. 

% English keywords
\end{abstract}

% The bottom paragraph of the front matter
\begin{bottompar}

% Related institue
\inst{Facultad de Ingeniería, Universidad Autónoma de Yucatán. Avenida de industrias no contaminantes y periférico norte s/n. Mérida, Yucatán, México.}

% Correspondence email of the correspondence author
\correspemail{carlosdotb@gmail.com}

% The journal's note
\journalnote{Este artículo de investigación es parte de Ingeniería–Revista Académica de la Facultad de Ingeniería, Universidad Autónoma de Yucatán, Vol. 22, No. 1, 2018, ISSN: 2448-8364}
\end{bottompar}
\end{frontmatter}
% End of front matter

% - - - - - - - - - - - - - - - - - - - - - - - - - - - - - - - - - - - - - - - - - - - - - - - - - - - - - - - - - - - - - - - - - - - - - - - - - %
% 											BEGIN WRITING YOUR PUBLICATION HERE													 %
% 											IN PURE LATEX 																							 %
% - - - - - - - - - - - - - - - - - - - - - - - - - - - - - - - - - - - - - - - - - - - - - - - - - - - - - - - - - - - - - - - - - - - - - - - - - %

\section{Introduction}
The problem of \textit{supervised learning} arises in contexts where a study of a dependent variable Y is necessary in terms of an independent variable X. The goal of supervised learning is to predict the values of Y given many instances of the variable X. Generally, we call the instances of X and the values Y takes, the \textit{inputs} and \textit{outputs}. In the pattern recognition context we typically may read \textit{features} and \textit{responses} respectively as an alternative terminology. With this in mind, it is noteworthy that we are in fact assuming the existence of a  function between X and Y such that Y = f(X). That is, there exists a relationship between the two variables. In general, we say there are two main tasks in supervised learning: \textit{classification} and \textit{regression}. We make the distinction between these two contexts by defining classification as the \textit{task that classifies inputs into several classes} and regression as the \textit{task which estimates the functional relationship between two variables}. To summarize the supervised learning concerns itself with the estimation of a function $\hat{f}$ such that $Y \approx \hat{f}(X)$. Admittedly, the interpretation of supervised learning is to let a mathematical or computational model ``learn by example''.  The examples being the values X takes, while the learning process is the identification of patterns given by the relationship between X and Y. Chiefly,  supervised learning is catalogued as a form of \textit{artificial intelligence} \cite{goodfellow2016deep}.

In recent years many supervised learning methods have been developed and many have been reviewed time and time again \cite{asingh, fernandezdel, kotsiantis, caruana}. One of the most recent reviews was conducted by Singh \textit{et al.} \cite{asingh}, which tests many different methods on a single dataset and presents the advantages and disadvantages in a textual manner and establishes their application areas. A slightly  older but extremely exhaustive analysis was conducted by Fernández-Delgado \textit{et al.} which evaluates 179 classifiers arising from 17 families on 121 datasets \cite{fernandezdel}. Support Vector Machines (SVB) and Random Forests are the top performing methods. Kotsiantis \textit{et al.} \cite{kotsiantis} have conducted a more comprehensive review of various different classification methods in which they show that SVM and Neural Networks (NN) were the best performing algorithms in terms of accuracy, classification speed and tolerance to parity problems. These algorithms, however, are lacking in terms of speed of learning, danger of overfitting and model parameter handling where Naïve Bayes and kNN are the top contenders. After analysing all of the results more carefully, one may argue that SVM clearly outperforms NN. This being said, one will find that NN and SVM are the most used supervised learning methods due to their simple ``black box'' functionality and so make good candidates to compare against. This work will not be concerned with such analyses but will instead propose a new supervised binary classification method. The method will be based in the resolution of a mathematical model through techniques from calculus of variations and functional analysis. Employment of these techniques allows us to exploit the underlying mathematical theory and properties from each of these fields. Even more, by utilizing these techniques we may obtain meaningful conditions in the continuous sense rather than performing optimization discretely. Comparing the performance of the model with SVM and NN we will determine if the new method is competitive or not.

\section{A Variational Approach for Supervised Learning}
\subsection{The Problem}
The supervised learning problem may be characterized as follows. Given a training set of $N$ observations

\begin{equation*}
\mathcal{T} = \{  (\mathbf{x}_1, y_1), \ldots, (\mathbf{x}_N, y_N) \  | \  \mathbf{x}_i \in \Omega, y_i \in  \mathcal{Y}\} ,
\end{equation*}
we would like to find a function $u = u(\mathbf{x})$ defined by $u: \Omega \rightarrow \mathcal{Y}$. In order to find a suitable function $u$ it is necessary to suggest a suitable model which best describes the problem. The most widely used framework to solve the supervised learning problem is the minimization of the sum of a loss function $L(u, y)$ with an added regularization term $S(u)$:
\begin{equation*}
\min_u \lambda S(u) + \sum^N_{i=1} L(u(\mathbf{x}_i), y_i) \ .
\end{equation*}
For the continuous case, however, we have:
\begin{equation} \label{eq:cont_problem}
\min_u \lambda S(u) + \int_\Omega L(u(\mathbf{x}_i), y_i) \ .
\end{equation}
The idea behind the model is to ``punish'' incorrect predictions using the loss function while the regularizer term allows us to ``control'' or ``regulate'' the variable $u = u(\mathbf{x})$. A regularization term is typically used to solve ill-posed problems or prevent overfitting. Thus, the regularization term should only depend on the variable being regularized; in our case $u$. We may further specialize $S(u)$ to our needs. Opting to punish the complexity of $u$, we might attempt to  ``measure'' the complexity and subsequently punishing it. Thereupon we let $R(u) > 0$ be a Lebesgue integrable function which measures the complexity of $u$. The exact definition of $R$ is left until a later section. This being said, we substitute back into Equation \ref{eq:cont_problem} to obtain:
\begin{equation} \label{eq:our_model}
\min_u \int_\Omega \Big[  L(u, y) + \lambda R(u)  \Big] d\mathbf{x} \ .
\end{equation}
This last equation will be our main interest for the remainder of this work.

One way to interpret Equation \ref{eq:our_model} is to notice that the \textit{empirical risk} of choosing $u$ among a space of \textit{hypothesis functions} $\mathcal{H}$ is defined as the expectation of the loss function $L$:

\begin{equation*}
\mathbf{E} \big[  L(u, y)  \big] = \int_\Omega L(u, y) d\mathbf{x}
\end{equation*}

Thus, if we wanted to minimize the empirical risk, it would amount to solving the following minimization problem:

\begin{align} \label{eq:emp_risk}
u^* &= \arg\min_{u \in \mathcal{H}} \mathbf{E} \big[  L(u, y)  \big]  \\
&= \arg\min_{u \in \mathcal{H}} \int_\Omega L(u, y) d\mathbf{x} \ ,
\end{align}
where $u^*$ is the \textit{minimizer} of the empirical risk.
Therefore, it is finally clear that we will be minimizing the \textit{regularized empirical risk for a choice of $u$}. Although Equation \ref{eq:emp_risk} is usually minimized through probabilistic and statistical methods, we will employ concepts from functional analysis and calculus of variations to find a minimization condition. The condition may then be solved numerically to obtain $u^*$.

\subsection{The Solution}
Akin to \textit{differential calculus}, we aim to ``differentiate'' the expression in Equation \ref{eq:our_model} with respect to $u$. Unfortunately, the usual definition of a ``derivative'' is insufficient for our purpose. Instead, we recall from functional analysis that if we find the functional derivative and equate to zero
\begin{equation*}
\dfrac{\delta F}{\delta u(\mathbf{x})} = 0
\end{equation*}
we obtain the \textit{optimality condition}. Solving this condition numerically will give us the minimizer $u^*$ of the functional $F[u]$. Before deriving the optimality condition, it is necessary to make the definition of our model more precise. Let $f$ be a function of the form:

\begin{align*}
f(\mathbf{x}, &u(\mathbf{x}), \nabla u(\mathbf{x})) = \\
&L(\mathbf{x}, u(\mathbf{x}))) + \lambda R(\mathbf{x}, u(\mathbf{x}), \nabla u(\mathbf{x})) \ ,
\end{align*}
with $y$ fixed \textit{s.t.} $L(u) = L(u, y)$. Furthermore, let $F[u]$ be the functional depending on $u$ \textit{s.t.}:
\begin{equation*}
F[u] = \int_\Omega f(\mathbf{x}, u(\mathbf{x}), \nabla u(\mathbf{x})) d\mathbf{x} \ .
\end{equation*}
Then, the functional derivative with respect to $u$ is given by the PDE Euler-Lagrange Equation:
\begin{align}
\dfrac{\delta F}{\delta u(\mathbf{x})} &= \dfrac{\partial f}{\partial u} - \nabla \cdot \dfrac{\partial f}{\partial \nabla u} \label{eq:t}
\\ &= \Big[ \dfrac{\partial L}{\partial u} +  \lambda \dfrac{\partial R}{\partial u}  \Big] - \nabla \cdot \Big[   \dfrac{\partial L}{\partial \nabla u} + \lambda \dfrac{\partial R}{\partial \nabla u}   \Big]
\\ &= \dfrac{dL}{du} + \lambda \dfrac{dR}{du} - \lambda \nabla \cdot \dfrac{\partial R}{\partial \nabla u}
\\ &= \dfrac{dL}{du} + \lambda (\dfrac{dR}{du} - \nabla \cdot \dfrac{\partial R}{\partial \nabla u}),
\end{align}
where $u = u(\mathbf{x})$ and $f = f(\mathbf{x}, u, \nabla u)$. The second line is obtained by substitution. The third line is obtained by remembering that $L$ does not explicitly depend on $\nabla u$ and thus it must be zero. The fourth line is the factorization of $\lambda$. After equating to zero, our original problem is then reduced to solving the following PDE Euler-Lagrange equation for particular choices of $L$ and $R$:
\begin{equation}\label{eq:general_eq}
\dfrac{dL}{du} + \lambda (\dfrac{dR}{du} - \nabla \cdot \dfrac{\partial R}{\partial \nabla u}) = 0 \ .
\end{equation}

\subsection{Radial Basis Function Approximation}
In order to approximate the function $u$, it is required we specify what kind of form it takes. Radial basis function (RBF) approximation relies on the idea that $u(\mathbf{x})$ and can be expressed as a weighted sum of radial basis functions $\{ \phi_i(\mathbf{x}) \}$, where the weights are the fixed parameter vector $\mathbf{w}$. In particular, the Gaussian RBF kernel is probably the most well known and the most widely used. For this reason only, it shall be our choice all along this work. Thus, bringing the previous ideas together, we write
\begin{equation*}
u(\mathbf{x}) = \sum^N_{i=1} w_i \phi(\mathbf{x}_i),
\end{equation*}
where  $\{ \phi_i(\mathbf{x}) \}$ will be a set of Gaussian RBF kernels given by
\begin{equation*}
\phi_i(\mathbf{x}) = e^{-c|| \mathbf{x} - \mathbf{x}_i ||^2 }
\end{equation*}

Maintaining the notation we have been using throughout this work, we say that the $\{ \mathbf{x}_i \}$ are the observations of the input variable $X$. In addition, $c$ is a positive constant that we are free to choose, while $|| \cdot ||$ is the Euclidean norm. We will call $c$ the \textit{fitting degree} of our model. As a small note, notice that by using our approximation of $u(\mathbf{x})$, we are setting the \textit{centers} of the RBFs as the observations. Intuitively, this will allow our model to make predictions by calculating the Euclidean distance between a point of which the class is known, and a new input which its class is unknown while assigning a ``weight'' to each point. This last step is akin to saying that some points will be more important than others when making the predictions.

\subsection{Cross Entropy}
In order to correctly employ a classification context we must give an appropriate loss function for classification. To find an appropiate loss, we take a Bernoulli variable $Y \in \{  0, 1 \}$. Our goal is then to predict the target class $y$ given an input $u(\mathbf{x}, \mathbf{w})$, where $\mathbf{w}$ is a vector parameter which $u$ depends on. From the previous section, we see that this is actually the case. To derive the loss function, we choose to maximize the likelihood that, given parameters $\mathbf{w}$, the model results in a prediction of the correct class for each input sample with the likelihood being a function of the parameters. In fact it is the minimization of the \textit{cross-entropy} \cite{crossentrop} (also called \textit{negative log-likelihood}) that effectively maximizes the likelihood. This function is given by:
\begin{equation*}
L(\sigma(u), y) =  - y \ln \sigma(u) - (1 - y) \ln \big(  1-  \sigma(u)  \big) . \label{eq:cross_entropy}
\end{equation*}
To use this function in our framework it is necessary to calculate its derivative with respect to $u$. Namely
\begin{equation*}
\dfrac{dL}{du} = \sigma(u) - y \ .
\end{equation*}

\paragraph{Notes.} This function is convex and so coupled with a convex regularization term our model will admit a unique solution $u^*$. Our prediction rule is $g(\sigma(u))$ where $g \in \{ 0, 1\}$ as opposed to $\sigma \in (0, 1)$. We outline that $\sigma$ gives us the probability of classifying an input $u$ as $y = 1$. To effectively ``binarize'' the output one must define $g$, for example, as $g = 1$ on $\sigma(u) \geq 0.5$ and $g = 0$ on $\sigma(u) < 0.5$.

\subsection{Laplacian Regularization $- \Delta u$} 
Now that we have derived an appropriate loss function for binary classification, we are tasked with the choice of a regularizer term. Much work has been done in the past to find a good regularizer and although there is no clear, best choice we can make do with most suggestions in the literature. In the recent work on this area, Belkin \textit{et al.} have proposed the square of the gradient vector norm, so that:
\begin{equation} \label{eq:gradient_norm_sq}
S(u) = \dfrac{1}{2} \int_\Omega || \nabla u ||^2 d\mathbf{x} \ ,
\end{equation}
for which $R(u, \nabla u) = || \nabla u ||^2$. The expression of the term depending on $R$ in Equation \ref{eq:general_eq} is calculated for this choice of $R$
:\begin{equation} \label{eq:laplacion_reg}
\dfrac{d}{du} \dfrac{||\nabla u||^2}{2}  - \nabla \cdot  \dfrac{\partial}{\partial \nabla u}  \dfrac{||\nabla u||^2}{2} = - \Delta u.
\end{equation}
One way to interpret this choice of regularization term is to regard $u(\mathbf{x})$ as a scalar field and to notice that taking the norm of a vector is interpreted as calculating its magnitude. Thus, this choice minimizes the slope of the gradient points in the direction of the greatest rate of increase. Therefore, it will prevent sharp edges and, in part, will ensure that $u^*$ be smooth and reduce overfitting.
\paragraph{Notes.}
Equation \ref{eq:gradient_norm_sq} is convex and thus there is a unique solution $u^*$ to our problem.
This equation is also known as the Dirichlet Energy functional \cite{evans}, and so we can have a more precise interpretation of minimizing this functional. Chiefly, the Dirichlet Energy measures how variable a function is and it is a quadratic functional on the Sobolev space $W^{k,2}$. In general, this regularization term has been tried with some favourable results by Lin \textit{et al.} \cite{tonglin} in the regression context.

\section{The Completed Model}
Without further ado, we shall give the whole model along with its solution. Let $u: \Omega \subset \mathbb{R}^m \rightarrow \mathbb{R}$ be the function we want to fit and let $y \in \{ 0, 1\}$ be the target values. Let $\sigma(u)$ be the probability that an input $u = u(\mathbf{x})$ is classified as $y = 1$. Let $F[u]$ be a functional depending on $u$, defined by

\begin{equation*}
F[u] = \int_\Omega -y \ln \sigma(u)  - (1 - y) \ln \big(  1-  \sigma(u)  \big)   +  ||  \nabla u  ||^2   \ d\mathbf{x}.
\end{equation*}
Then the condition of optimality is the elliptic PDE 
\begin{equation*}
\sigma(u) - y - \Delta u = 0 \ .
\end{equation*}
Here, our equation takes on a more special interpretation. Isolating the variable $y$ on the left hand side, we see that actual form that $y= y(\mathbf{x})$ must take is that of the right hand side. Therefore, this expression is the function used to predict the outputs.
\begin{equation*}
y(\mathbf{x}) = \sigma(u(\mathbf{x})) - \Delta u(\mathbf{x}) \ .
\end{equation*}
In order to solve the problem numerically, we can isolate the variable $y$ to see that we have reduced the original problem to another fitting problem that can be solved numerically. Recalling that $u = u(\mathbf{x})$ depends on a fixed weight vector $\mathbf{w}$ and can be expressed by the sum of the set of RBFs $\{ \phi_i(\mathbf{x}) \}$, it is required to find the weight vector $\mathbf{w}^*$ which appropriately satisfies Equation \ref{eq:subproblem_lr}.
\begin{equation} \label{eq:subproblem_lr}
\underbrace{ \sigma(u) - \Delta u}_{\text{Function to fit}}   = \underbrace{y}_{\text{Targets}}
\end{equation}
Since we have $N$ datum pairs $(\mathbf{x}_i, y_i)$ we may attempt to utilize these in order to find the function, or equivalently the weights $\mathbf{w}^*$, which satisfy the equality above. It is then that we may write the problem of finding $\mathbf{w}^*$ as a least squares problem (LSQP). Let $g(\mathbf{x}_i, \mathbf{w}) = \sigma(u(\mathbf{x}_i, \mathbf{w})) - \Delta u(\mathbf{x}_i, \mathbf{w})$, then the LSQP is:
\begin{align}
\mathbf{w}^* &= \arg \min_{\mathbf{w}} \sum^N_{i=1} \big[   y_i   +  \Delta u(\mathbf{x}_i, \mathbf{w})  -  \sigma(u(\mathbf{x}_i, \mathbf{w}))    \big]^2 \\
&= \arg \min_{\mathbf{w}} \sum^N_{i=1} \big[ y_i - g(\mathbf{x}_i, \mathbf{w})  \big]^2  \\
&=  \arg \min_{\mathbf{w}} \sum^N_{i=1} \big[ y_i - g_i(\mathbf{w})  \big]^2 \\
&= \arg \min_\mathbf{w} S(\mathbf{w}) \label{eq:least_squares},
\end{align}
Note we have defined $g_i(\mathbf{w}) = g(\mathbf{x}_i, \mathbf{w})$ to obtain the third step. The problem is therefore reduce to solving this problem. In our particular case, we have chosen to solve this problem via Levenberg-Marquardt \cite{levenberg}; a widely used numerical algorithm used to solve non-linear least squares problems. A detailed description of this algorithm can be found in  the book \textit{Numerical Recipes in C} \cite{williamnum}. Using this algorithm will mean that our model will depend on yet another parameter; $\eta$ is a dampening parameter which will allow to solve a ``dampened'' version of the problem above.

\section{Evaluation Methodology}
We tested our model against 9 binary datasets. For each dataset we calculated two metrics:
\begin{itemize}
\item Accuracy
\item Area under the ROC curve (AUC) \cite{hanleyroc}
\end{itemize}
Both of the metrics are calculated over the test set in a  repeated 5-\textit{fold} cross validation scheme. There has been minimal preprocessing of the datasets. That is, we have centered to 0 mean and unit variance the features of each dataset. Namely, we have \textit{standarized} the dataset. Notably, no dimensionality reduction has been applied. 

When training the models, it is necessary to specify the parameters in which they depend on. For our model (LR) we have two choose a parameter triplet $(c, \lambda, \eta)$; $c$ is the fitting degree, $\lambda$ is the regularization parameter and $\eta$ is a dampening parameter. For a \textit{RBF-kernel} SVM, we choose the parameter pair $(C, \gamma)$; $C$ is the penalty parameter of the error term while $\gamma$ is the kernel coefficient. Lastly, we assume that NN is a \textit{multi-layer perceptron} with one hidden layer of 100 nodes which depends on a  regularization term $\alpha$. Both SVM and NN are implementations of the Python library \texttt{sklearn} \cite{sklearn}. The parameters for each of our models were chosen by the following methodology:
\paragraph{\textbf{LR.}}
The parameter triplet $(c, \lambda, \eta)$ is searched on a grid:
\begin{itemize}
\item $c$ is searched on the interval $(0, 5)$ with step $ = \ln 2$ .
 \item $\lambda$ is searched on the interval $[0, 10]$ with step $= 1$ .
 \item $\eta$ is fixed to $\eta = 1$ for each and every dataset.
\end{itemize}
The value of $\eta = 1$ was found empirically to work well with almost any value of $\lambda$ and $c$. To threshold the outputs of LR into values of the set $\{ 0, 1\}$ we will use the following classification rule:
\begin{equation*}
g(\mathbf{x}) = \begin{cases}
0 & \sigma(u(\mathbf{x})) - \Delta u(\mathbf{x}) \leq \dfrac{1}{2} \vspace{2mm} \\
1 & \sigma(u(\mathbf{x})) - \Delta u(\mathbf{x}) > \dfrac{1}{2}
\end{cases}
\end{equation*}

\paragraph{\textbf{SVM.}}
The parameter pair $(C, \gamma)$ is searched on a grid:
\begin{itemize}
\item $C$ is searched on the interval $(0, 5)$ with step $ = \ln 2$ .
\item $\gamma$ is fixed to $\gamma = 1 / m$, where $m$ is the number of features of the dataset.
\end{itemize}
Originally $\gamma$ was searched on the interval $[0, 10]$ with step $= 1$. It was soon found empirically, that fixing $\gamma$ to $\gamma = 1 / m$ resulted in better performance.

\paragraph{\textbf{NN.}}
The parameter $\alpha$ was fixed to $\alpha = 0.001$.

\section{Results and Analysis}
This section presents the results in Tables \ref{tab:acc} through \ref{tab:grades}. To understand more easily the performance of the methods, we have arranged the results into various tables. In terms of accuracy,  LR outperformed SVM and NN on 3 datasets. In detail, LR outperformed NN on 4 datasets and outperformed SVM on 2, tying on the Breast Cancer dataset. To fully grasp how much better or worse our method has performed we calculated the absolute value of the residual between LR and the top performer for each dataset. On average we see that our method was down by $0.9567$ \%. Although not shown in Table \ref{tab:acc}, the average distance from the top performer for SVM is $0.5229$ \% and for NN it is $0.8975$ \%. These scores lend to the interpretation of ``which method got closer to the real solution''. Looking at who got the top score we might say that the top performer was the best \textit{method suited for that particular type of dataset}. It is of interest then to evaluate a method which in average should perform well for most types of datasets. Interpreting the results, we can confidently say that \textbf{SVM is the best method, while NN is second and LR comes a close third in terms of accuracy}.

% Use \begin{table*} \center .... \end{table*}
% To make the table go down the center
\begin{table*}[h!]
\center
\caption{The accuracy (\%) of each method is outlined in this table. The second to last column indicates which ranking (1st, 2nd or 3rd) LR obtained; higher is better. On the other hand, the last column indicates the absolute value of the residual between LR and the 1st place. In bold, the best accuracy.} \vspace{3mm}
\setlength{\tabcolsep}{2mm}
\begin{tabular}{c|cc|ccc|cc}
\toprule
Data & Dim & $N$ & LR & SVM & NN & Place & Dist. from 1st \\
\midrule
\midrule
Australian 				& 14 	& 690  	& 86.6667 					& 86.8116 						& \textbf{87.8261} 	& 3rd 					&  1.1594 \\
Blood Transfusion 	& 4 		& 748 		& \textbf{78.2246} 	& 78.2237 						& 77.2859 					& 1st				 	& 0.0\\
Breast Cancer 			& 30 	& 569 		& 97.7146 					& \textbf{98.7425} 		& 98.2673 					& 3rd 					& 1.0279\\
Bupa 						& 6 		& 345 		& \textbf{72.4638} 	& \textbf{72.4638 }		& 71.0145 					& 1st 					& 0.0\\
German 					& 24 	& 1000 	& 76.0000 					& 76.6000 						& \textbf{78.3000} 	& 3rd 					& 2.3 \\
Haberman 				& 3 		& 306 		& 73.5431 					& 73.8710 						& \textbf{74.5267} 	& 3rd 					& 0.9836 \\
Heart 						& 13 	& 270 		& 82.2222 					& \textbf{84.8741}			& 84.0148 					& 3rd 					& 2.6519 \\
Sonar  						& 60 	& 208 		& 88.4321 					& \textbf{88.9199} 		& 87.4681 					& 2nd 					& 0.4878\\
Vertebral Column 	& 6 		& 310 		& \textbf{86.7742} 	& 85.4839 						& 83.8710 					& 1st 					& 0.0 \\
\midrule
 & & & & \multicolumn{3}{r}{Average distance from 1st:} 		& 0.9567 \\
\bottomrule
\end{tabular}
\label{tab:acc}
\end{table*}

The AUC score may also be used to further determine the performance of a classifier \cite{fawcett}. Proceeding in a similar manner as before, we calculate the average residual between the 1st place and LR. We find that LR is on average down from $0.0126$ units from the top performer for each dataset. The same is calculated for SVM and NN; respectively $0.0088$ and $0.0161$. Surprisingly, even though NN outperformed SVM and LR on 4 different datasets while LR only outperformed the others on one, \textbf{on average LR will perform better than NN.} Unsurprisingly, SVM will still perform better than NN and LR. 

\begin{table*}[h!]
\center
\setlength{\tabcolsep}{2mm}
\caption{The area under the ROC curve has been calculated for each method over each dataset Also, we have added a grade next to the scores to see how they perform against each other more easily; A being excellent performance, while F is catalogued as a fail. Likewise (Table \ref{tab:acc}), the second to last column indicates which ranking (1st, 2nd or 3rd) LR obtained; higher is better. On the other hand, the last column indicates the absolute value of the residual between LR and the 1st place. In bold, the best AUC.} \vspace{3mm}
\begin{tabular}{c|ccc|cc}
\toprule
Data & LR & SVM & NN & Place & Dist. from 1st \\
\midrule
\midrule
Australian 					& 0.8643  (B)				& 0.8701(B)					& \textbf{0.8777} (B)		& 3rd 					& 0.0134\\
Blood Transfusion 		& 0.5844  (F)				& \textbf{0.6144} (D)		& 0.5502	 (F)					& 2nd				 	& 0.03 \\
Breast Cancer 				& 0.9729  (A)				& 0.9800 (A)					& \textbf{0.9858} (A) 	& 3rd 					& 0.0129 \\
Bupa 							& 0.7024  (C)				& \textbf{0.7059} (C)		& 0.6866	(D)					& 2nd				 	& 0.0035 \\
German 						& 0.6821	  (D)				& 0.6920	(D)					& \textbf{0.7058} (C)		& 3rd 					& 0.0237 \\
Haberman 					& \textbf{0.5560} (F)	& 0.5559 (F)					& 0.5463 (F)					& 1st 					& 0.0 \\
Heart 							& 0.8180  (B)				& 0.8452	(B)					& \textbf{0.8322} (B)		& 3rd 					& 0.0142 \\
Sonar  							& 0.8857  (B)				& \textbf{0.8906} (B)		& 0.8812 (B)					& 2nd 					& 0.0049 \\
Vertebral Column 		& 0.8292  (B)				& \textbf{0.8404} (B)		& 0.7978	 (C)					& 2nd 					& 0.0112 \\
\midrule
& & & \multicolumn{2}{r}{Average distance from 1st: } & 0.0126 \\
\bottomrule
\end{tabular}
\label{tab:roc}
\end{table*}

Even though our analysis has been very exact until now, it is time to present a more intuitive analysis based on the AUC. Furthermore, this analysis is more \textit{robust} than the previous. The analysis consists in assigning a ``grade'' to a classifier by specifying the following grading scheme:
\[
\text{Grade} = 
\begin{cases}
\text{Excellent} \ (A) & 0.9 \leq \text{AUC}  \leq 1 \\
\text{Good} \ (B) & 0.8 \leq \text{AUC} < 0.9 \\
\text{Fair} \ (C) & 0.7 \leq \text{AUC} < 0.8 \\
\text{Poor} \  (D) & 0.6 \leq \text{AUC} < 0.7 \\
\text{Fail} \ (F) & 0.5 \leq \text{AUC} < 0.6 \\
\end{cases}
\]
The ``robustness'' comes from the fact that we are partioning discretely the interval $[0, 1]$ and assigning each a grade. Small variations within the sub-intervals will be neglected. Looking at Table \ref{tab:roc} we see that each of the AUC scores has a letter assigned to it. This is interpreted as the grade which the method received on that particular dataset. In order to summarize the grades, we arrange the number of times a method received a particular grade in Table \ref{tab:grades}. Simply by looking at Tables \ref{tab:roc} and \ref{tab:grades} we can get the sense that all of the models performed similarly. In order to obtain a quantitative measure, we may assign each grade a value. Namely, A = 1, B = 2, $\ldots$, F = 5. Obtaining the weighted total of the grades will then let us asses directly which method is better by looking at the lowest total. The results are presented in Table \ref{tab:grades}. Immediately, we see that SVM once again obtained the best score. This time, however, LR came a close second  with only a 1 point difference. On the contrary NN was down by 4 points from SVM and 3 points from LR. Summarizing, \textbf{LR outperforms NN again while SVM remains overall the best method.}

\begin{table*}[h!]
\center
\caption{The columns LR, SVM and NN indicate the numbers of times each of the methods got the grade on the leftmost column. Letting A = 1, B = 2, C = 3, D = 4, F = 5, we can calculate a weighted final grade for our classifiers and see how each of them performed. Clearly a lower grade is better. The weighted grade is just the weighted total of the grades each of these methods obtained.}
\vspace{3mm}
\setlength{\tabcolsep}{2mm}
\begin{tabular}{c|ccc}
\toprule
Grade & LR & SVM & NN \\
\midrule
A & 1 & 1 & 1\\
B & 4 & 4 & 3 \\
C & 1 & 1 & 2 \\
D & 1 & 2 & 1 \\
F & 2 & 1 & 2 \\
\midrule
Weighted grade (lower is better): & 26 & \textbf{25} & 29\\
\bottomrule
\end{tabular}
\label{tab:grades}
\end{table*}

\section{Conclusion}
Even though supervised learning has had tremendous advances in the last few years, it remains clear that a lot of work has yet to be done. 
As part of our contribution to the field, we have designed and implemented a new variational model for binary classification. Opting to attack the supervised learning problem by functional and variational means, we arrived at the LR model and gave each of its components interpretations. Moreover we used the Levenberg-Marquardt method to solve the optimality condition which in essence was equivalent to ``training the model''. We outlined our evaluation  criteria and methodology so that we could compare how well our method stood against two of the most utilized methods; NN and SVM.  From the results obtained, we conducted three different analyses to quantify the performance of each model. The accuracy results obtained, clearly indicated that SVM was the superior method amongst the three, and so the focus shifted to the comparison between NN and LR. It was found that NN outperformed LR by a small margin. The second analysis showed that LR outperformed NN while SVM came out on top once again. Finally, we conducted a more robust analysis which permitted us to ignore small variations in the scores. This last analysis further showed that LR clearly outperformed NN. Moreover it showed that LR was not far from SVM's performance. We speculate that this is due to the heavy optimizations in the implementation of SVM which permit it to find the minimum in a faster manner.

Without a doubt, the variational approach looks to be very promising and is to be explored in future work. Namely, we will concentrate in deriving expressions for higher order derivatives and better formalizing the problem at hand in order to derive more interesting properties.
The implementations for the LR model, benchmarking scripts and datasets may all be found at \texttt{https://github.com/carlosb/thesis}.

% Needed for converting the bibliography back to one colum
\onecolumn 

% The bibliography

\end{document}